\documentclass[conference]{IEEEtran}
\IEEEoverridecommandlockouts

\usepackage{cite}
\usepackage{amsmath,amssymb,amsfonts}
\DeclareMathOperator*{\argmax}{arg\,max}
\usepackage{algorithmic}
\usepackage{graphicx}
\usepackage{textcomp}
\usepackage{xcolor}
\usepackage{enumitem}
\usepackage{multirow}
\usepackage{booktabs}
\def\BibTeX{{\rm B\kern-.05em{\sc i\kern-.025em b}\kern-.08em
    T\kern-.1667em\lower.7ex\hbox{E}\kern-.125emX}}
\begin{document}

\title{
	An Event-Driven Framework for Fly-Inspired Visual Motion Detection
	\thanks{This research was supported by the National Natural Science Foundation of China under grant nos. 62376063, 12571558, 12571491.}
}

\author{
	\IEEEauthorblockN{Qinbing Fu$^{*,\dagger,1}$, Jingyu Huang$^{\dagger,1}$, Yan Xie$^{1}$, Jigen Peng$^{1}$, Yuchao Tang$^{*,1}$}
	\IEEEauthorblockA{$^{1}$ School of Mathematics and Information Science, Guangzhou University, China}
	\IEEEauthorblockA{$^{*}$ Corresponding authors: \{qifu,yctang\}@gzhu.edu.cn}
	\IEEEauthorblockA{$\dagger$ The authors contributed equally.}
}

\maketitle

\begin{abstract}

Fast and reliable motion detection is essential for machine vision and autonomous systems operating in dynamic environments. 
This work integrates emerging event-based sensing with biologically structured neural computation to establish an efficient computational paradigm for visual motion detection. 
The proposed framework is built upon a recently developed fly-inspired neural network that emulates motion-processing circuits in the optic lobe. 
Owing to its feed-forward and training-free architecture, the neural model requires only a small number of interpretable parameters and is well suited for real-time implementation. 
Event cameras provide low-latency, low-power, and high-dynamic-range visual sensing by asynchronously transmitting brightness-change events. 
However, their performance can be degraded by event noise, including temporal noise and junction-leakage-induced activity, particularly under low-light conditions. 
Moreover, effective integration between event-based visual representations and biologically inspired neural processing remains under-explored. 
To address these challenges, we propose an event-driven computational framework that combines time-surface encoding for front-end event representation with a fly optic-lobe-inspired neural network for foreground motion-direction estimation. 
A bottom-up attention mechanism is further incorporated to suppress background motion and enhance the saliency of foreground targets. 
The proposed method is evaluated via real-world datasets of ground-vehicle detection and compared with a baseline frame-based model and an optimization-based approach. 
Experimental results demonstrate that the framework effectively combines the temporal advantages of event-driven vision with the efficiency and interpretability of bio-inspired neural processing. 
These findings support a new computational paradigm for integrating event cameras with structured neural models for motion detection.

\end{abstract}


\section{Introduction}
\label{Sec: Introduction}

Visual motion detection is a fundamental capability of both biological and artificial vision systems. 
It provides essential information for autonomous navigation, collision avoidance, target tracking, and swarm coordination in dynamic environments. 
Consequently, developing efficient and reliable motion-detection methods remains an important research challenge.

Recently, event cameras have emerged as a promising bio-inspired sensing paradigm for dynamic vision tasks. 
Owing to their asynchronous operation, event cameras offer microsecond-level temporal resolution, low latency, and reduced motion blur, making them particularly well suited to motion analysis. 
Accordingly, event-based vision has been widely explored for applications such as optical flow (OF) estimation, motion segmentation, and visual navigation, and so forth as surveyed in \cite{event-survey}. 
For example, \cite{seg2019} proposed an event-wise motion-compensation method for motion segmentation, jointly estimating event-cluster assignments and motion parameters to separate independently moving objects from the background. 
For OF estimation, \cite{max2025} incorporated edge information into a contrast-maximization framework, resulting in an edge-informed approach for event-based flow estimation. 
In addition, \cite{track2015} addressed real-time object tracking using an event-driven shape-tracking algorithm.

Inspired by the efficiency of biological visual systems, \cite{brosch2015} investigated event-based OF estimation from a biological perspective and proposed a motion-detection framework based on spatiotemporal receptive fields and response-normalization mechanisms. 
This approach was motivated by the directional selectivity of neurons in the primary visual cortex (V1). 
In addition, \cite{orchard2013} proposed a spiking neural network architecture capable of estimating motion direction and velocity directly from asynchronous events, thereby exploiting the spike-based nature shared by biological neural signaling and event streams. 
Subsequently, \cite{IBM2018} developed a fully spike-based OF network inspired by a neural mechanism observed in the rabbit retina and deployed it on neuromorphic hardware, enabling real-time motion estimation with extremely low power consumption. 
More recently, \cite{unsuper2020} introduced a hierarchical spiking neural network that learns local and global motion representations from event streams in an unsupervised manner, further improving the biological plausibility of event-based motion-perception models.

Despite these advances, several challenges remain. 
Current event-based OF methods often rely on computationally intensive optimization procedures or dense spatiotemporal operations, which can limit real-time performance. 
Although spiking neural network models offer biologically plausible solutions, they commonly require large-scale architectures, specialized neuromorphic hardware, or lengthy training procedures, increasing implementation complexity and deployment cost. 
Therefore, developing a lightweight and biologically plausible motion-perception framework capable of efficiently processing event streams remains an important research challenge.

To address these challenges, we propose a lightweight computational framework inspired by motion processing in the stratified neuropil layers of the insect optic lobe. 
Several insect-inspired models have been well established and have demonstrated robust motion-detection capability across a range of complex and dynamic real-world scenarios, e.g., \cite{fu2020,fu2026}. 
However, these approaches are primarily designed for frame-based cameras and can be adversely affected by low-light conditions and highly dynamic background interference. 
Moreover, they have not been systematically integrated with event-based sensing. 
This raises an important question: can the complementary advantages of event cameras and fly-inspired visual processing be integrated into an efficient framework for robust motion detection in challenging real-world environments?

We answer this question affirmatively by proposing a bio-inspired motion-perception framework that integrates event-stream encoding with a representative fly optic-lobe neural processing architecture adapted from \cite{fu2020}. 
As the front-end module, a histogram-of-averaged-time-surfaces (HATS) representation is employed to encode polarity-specific ON/OFF event streams. 
This mechanism synchronizes asynchronous events within a selected temporal window and combines exponential time decay with spatial averaging and downsampling to construct compact spatiotemporal representations resembling short-term visual memory. 
The resulting ON/OFF event representations are then processed by a stratified neural network to estimate the motion direction of foreground objects. 
Furthermore, a bottom-up attention mechanism is incorporated to dynamically highlight salient motion regions, thereby improving target-selective motion detection against complex natural backgrounds.

\begin{figure}[t]
	\centering
	\vspace{-20pt}
	\includegraphics[width=0.4\textwidth]{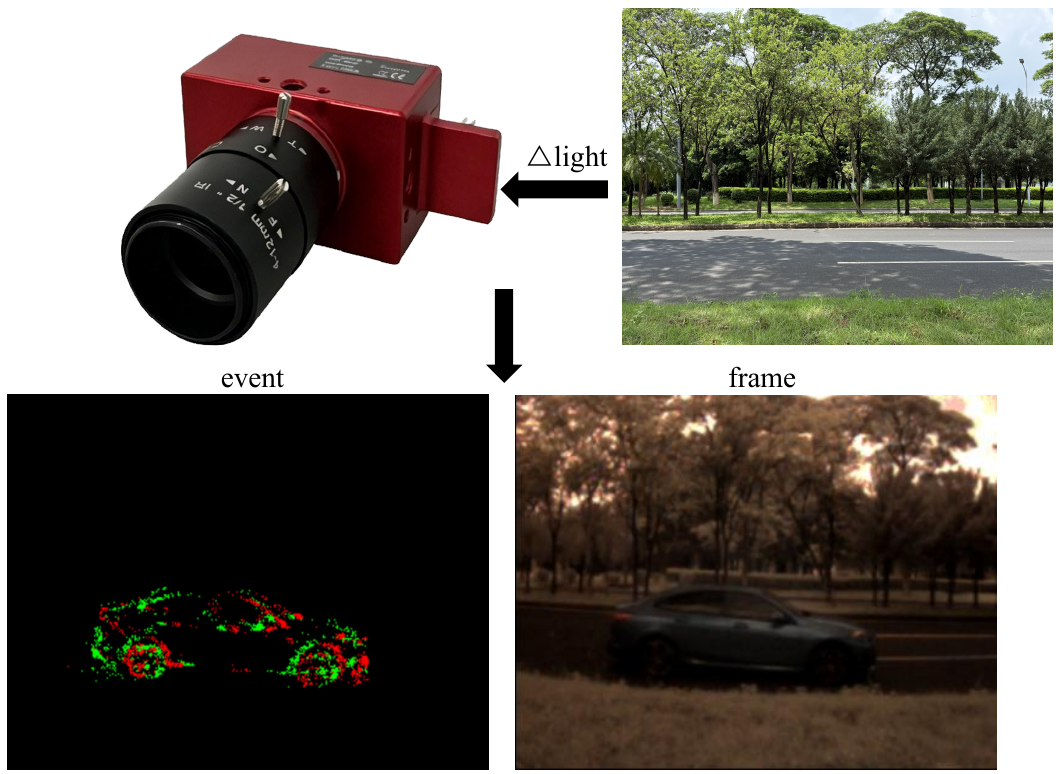}
	\caption{
		Schematic operating principle of the DAVIS camera: pixel-level brightness changes asynchronously trigger ON/OFF events (red/green), while active pixel sensor (APS) intensity frames are captured and output at a fixed frame rate.
		}
	\label{fig:davis}
	\vspace{-10pt}
\end{figure}

Systematic experiments were conducted to evaluate the effectiveness and robustness of the proposed method using real-world ground-vehicle sequences captured with a DAVIS event camera (Fig.~\ref{fig:davis}). 
Its performance was compared with that of the baseline frame-based model \cite{fu2020} and an event-driven optimization-based approach \cite{contrast2019}. 
The results demonstrate that the proposed computational framework can stably extract motion features, including motion direction and velocity-related responses, in both daytime and low-light nighttime scenarios. 
Meanwhile, it maintains low computational complexity with a small scale of tractable parameters, and a biologically plausible processing structure.


\section{Computational Framework}
\label{Sec: method}

\subsection{Event Camera}

The proposed framework processes data streams acquired using a DAVIS346 event camera \cite{event-survey}. 
As illustrated in Fig.~\ref{fig:davis}, the DAVIS sensor simultaneously outputs asynchronous events and synchronous intensity frames from a single imaging device. 
This hybrid sensing mechanism enables high-temporal-resolution event acquisition while preserving conventional image recordings, making it well suited for motion-perception research.

The DAVIS346 sensor provides a spatial resolution of 346 $\times$ 260 pixels for both event and frame outputs. 
Its event stream offers microsecond-level temporal resolution, latency below 1~ms, a dynamic range of up to 120~dB, and a maximum throughput of 12 million events per second. 
In contrast, the APS frame output operates at up to 40~FPS with a dynamic range of approximately 55~dB. 
However, its effective frame rate can decrease substantially under low-light conditions, falling below 10~FPS in our recordings. 
These complementary sensing characteristics enable the DAVIS camera to capture rapid temporal variations while retaining conventional image appearance information.

Mathematically, the event stream can be represented as a discrete set of events:
\begin{equation}
	\mathcal{E} = \{e_i\}_{i=1}^I, \quad e_i = (\mathbf{x}_i, t_i, p_i),
\end{equation}
where $\mathbf{x}_i = (x_i, y_i)$ denotes the pixel coordinates that triggered the event, $t_i \geq 0$ is the timestamp at which the event was generated. 
$p_i \in \{-1, 1\}$ indicates the event's polarity, where $-1$ and $1$ correspond to OFF and ON events, respectively.

\begin{figure*}[t]                
	\centering
	\vspace{-20pt}
	\includegraphics[width=0.95\textwidth]{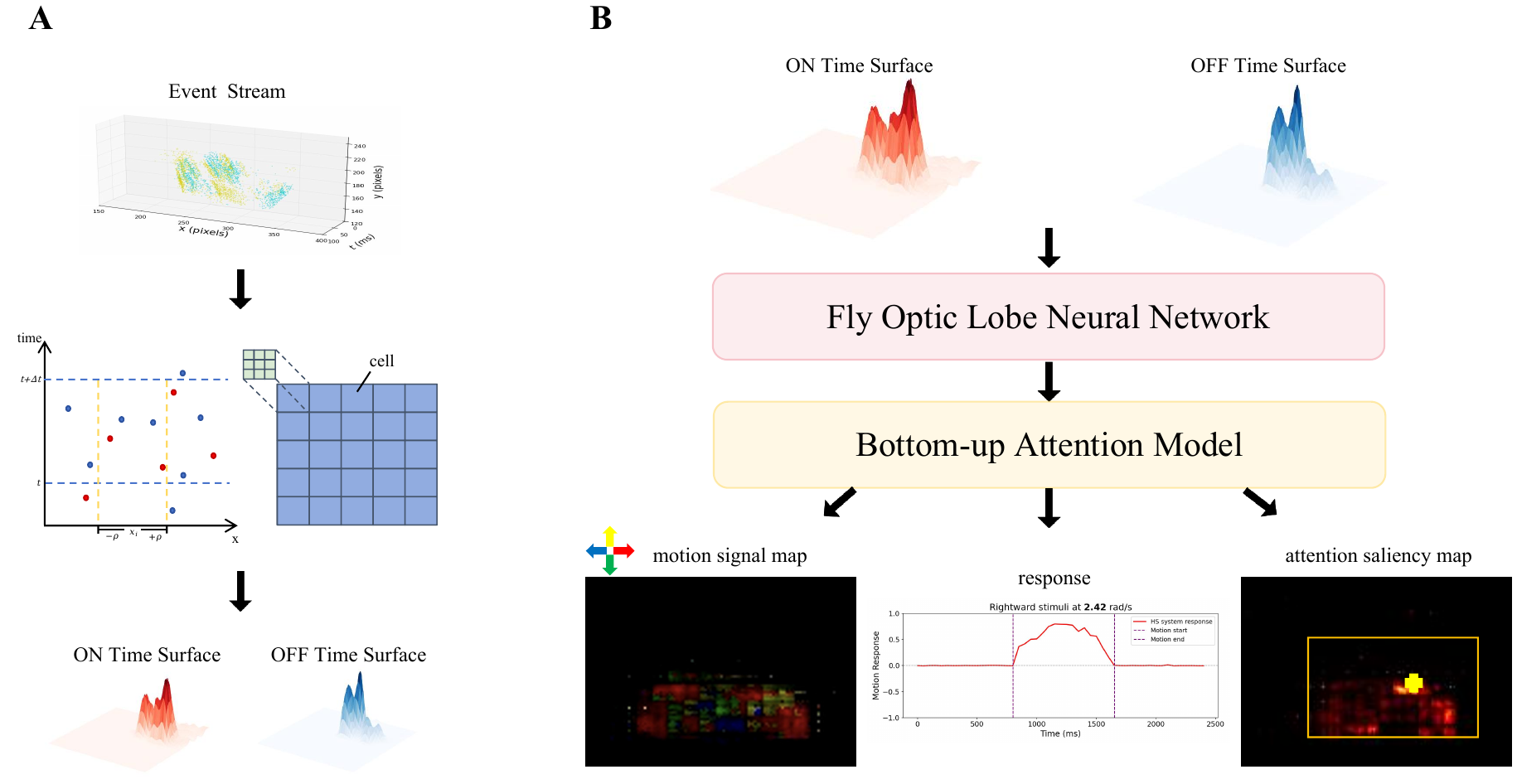}
	\caption{
		Overview of the proposed computational framework. 
		Panel A: The input ON/OFF event streams are first encoded into HATS representations. 
		Panel B: The two polarity-specific time surfaces are fed into a fly optic-lobe-inspired neural network to extract directional motion signals from the foreground moving target, producing motion responses that indicate both motion direction and magnitude. 
		A saliency map is further generated through a bottom-up attention model, which produces a binary mask centered on the estimated visual centroid to suppress isolated background noise and irrelevant motion responses.
	}
	\label{fig:flow}
	\vspace{-10pt}
\end{figure*}

\subsection{Events Representation}

We represent polarity-specific events using time surfaces encoding method (Fig.~\ref{fig:flow}). 
We compute the time surfaces using the history of events within a temporal window of size $\Delta t$ to characterize the local spatiotemporal patterns surrounding an event. 
Specifically, we define a local memory time surface $\mathcal{T}_{e_i}$ as follows:
\begin{equation}
	\mathcal{T}_{e_i}(\mathbf{z}, q) = 
	\begin{cases} 
		\sum_{e_j \in \mathcal{N}_{(\mathbf{z}, q)}(e_i)} e^{-\frac{t_i - t_j}{\tau}} & \text{if } p_i = q, \\
		0 & \text{otherwise},
	\end{cases}
	\label{Eq:ts}
\end{equation}
where
\begin{equation}
	\mathcal{N}_{(\mathbf{z}, q)}(e_i) = \{e_j : \mathbf{x}_j = \mathbf{x}_i + \mathbf{z}, \ t_j \in [t_i - \Delta t, t_i), \ p_j = q\}.
	\label{Eq:tw}
\end{equation}
The polarity $q \in \{-1, 1\}$ represents OFF and ON events respectively. 
$\mathbf{x}_i + \mathbf{z}$ indicates a spatial neighborhood centered by $\mathbf{x}_i$, and $\tau$ is a decay factor that assigns smaller weights to events farther in the past, the events that are closer together have greater weight.

We divide the image into cells of size $K\times K$ in space. 
For each cell $C$ (3$\times$3 in image pixels), we accumulate the time surfaces of all events within $C$ into a histogram. 
Specifically, the calculation is
\begin{equation}
	\mathbf{h}_C(\mathbf{z}, p) = \sum_{e_i \in C} \mathcal{T}_{e_i}(\mathbf{z}, p),
\end{equation}
where  $e_i \in C$  indicate that the pixel-wise events fall within the cell $C$.

Due to the inherent characteristics of event-based sensors, they are relatively sensitive to object contrast. 
Objects with higher contrast trigger more events, while those with lower contrast produce fewer. 
To mitigate the impact of contrast variations on the results, we compute the average over the total number of events within the temporal window, yielding the averaged histogram:
\begin{equation}
	\bar{\mathbf{h}}_C(\mathbf{z}, p) = \frac{1}{|C|} \sum_{e_i \in C} \mathcal{T}_{e_i}(\mathbf{z}, p). 
\end{equation}

Notably, to simulate the ON and OFF pathways in the biological visual system \cite{ziyan2025}, the generated time surfaces are further decomposed according to the event polarity. 
For positive events, we have
\begin{equation}
	\mathbf{h}_C^{on}(\mathbf{z}) = \bar{\mathbf{h}}_C(\mathbf{z}, p),\ \text{s.t.}\ C = \{e_i \in C \mid p_i = +1\}.
\end{equation}
For negative events, we have
\begin{equation}
	\mathbf{h}_C^{off}(\mathbf{z}) = \bar{\mathbf{h}}_C(\mathbf{z}, p),\ \text{s.t.}\ C = \{e_i \in C \mid p_i = -1\}.
\end{equation}

Subsequently, the histograms of all cells are reassembled to yield the positive and negative time surfaces, respectively:
\begin{equation}
	\mathcal{H}_{on/off} = \mathcal{A}(\mathbf{h}_{C_1}^{on/off}, \mathbf{h}_{C_2}^{on/off}, \dots, \mathbf{h}_{C_K}^{on/off}),
\end{equation}
where $\mathcal{A}(\cdot)$ represents the process of arranging each unit histogram according to the spatial position of its corresponding cell in the original image.

\subsection{Fly Optic Lobe Network}

The original baseline model was designed for frame-based visual processing of discrete digital signals and formulated as a visual neural network that mimics motion processing in the fly optic lobe \cite{fu2020}. 
Building upon the ON/OFF event representation, we feed the polarity-specific time surfaces into the baseline model in place of conventional brightness-based ON/OFF contrast signals, thereby forming the proposed network. 
Specifically, $\mathcal{H}_{on}$ and $\mathcal{H}_{off}$ are provided to the ON and OFF channels of the motion-perception network, respectively. 
This design preserves the temporal information encoded in the event stream while maintaining compatibility with the biologically inspired motion-processing architecture.

For brevity, the motion pre-filtering and ON/OFF pathway separation stages are not reiterated here, as they follow the algorithms presented in \cite{fu2020}. 
Instead, we briefly describe the T4- and T5-inspired motion-correlation stage, in which an elementary motion detector (EMD) structure integrates spatially offset signals to generate selective responses to motion along the four cardinal directions $\mathcal{D} \in \{r, l, u, d\}$.

The T4/T5 interneurons generate direction-selective responses by correlating the delayed signals at adjacent positions in the parallel ON/OFF pathway with the non-delayed signals, respectively. 
Taking the preferred direction of rightward movement as an example, this calculation can be expressed as
\begin{equation}
	\begin{aligned}
		T4_r(x,y,t) &= \sum_{i = sd}^{sd \cdot n_c} \hat{\mathcal{H}}_{on}(x,y,t) \cdot \mathcal{H}_{on}(x+i,y,t),\\
		T5_r(x,y,t) &= \sum_{i = sd}^{sd \cdot n_c} \hat{\mathcal{H}}_{off}(x,y,t) \cdot \mathcal{H}_{off}(x+i,y,t),\\
	\end{aligned}
	\label{Eq:sd}
\end{equation}
where $\hat{\mathcal{H}}$ denotes the temporally delayed signal, $sd$ and $n_c$ indicates spatial distance and number of spatially connected correlation neurons, respectively. 
Similar correlation operations are employed to detect leftward, upward, and downward motions.

\begin{table}[t]
	\centering
	\vspace{-20pt}
	\caption{
		\small
		Parameters of Model
	}
	\label{Tab-Parameters}
	\begin{tabular}{c|c|c}
		\hline
		\toprule
		Parameter	& Description	& Value \\
		\midrule
		$\tau$	& time constant in time surface (Eq.~\ref{Eq:ts})	& 30ms\\
		$\Delta_t$	& temporal window in time surface (Eq.~\ref{Eq:tw})	& 50ms\\
		$sd \times n_c$	& spatial correlation distance (Eq.~\ref{Eq:sd})	& 4$\times$4\\
		$\{g_m, g_p\}$	& gain factors in steering response (Eq.~\ref{Eq:sr})	& $\{\text{5}, \text{30}\}$\\
		$R$	& search side length (Eq. \ref{Eq:roi})	& 50\\
		\bottomrule
	\end{tabular}
	\vspace{-10pt}
\end{table}

\subsection{Bottom-Up Attention}

Inspired by selective-attention mechanisms in biological visual systems, a bottom-up attention module is introduced to enhance target-selective foreground motion perception. 
Rather than processing the entire visual field uniformly, the proposed mechanism dynamically prioritizes salient motion regions, thereby suppressing irrelevant background activity and isolated noise responses.

The attention-driven salience map highlights the region of interest showing the strongest motion response. 
We select the pixel coordinates with the maximum value in the salience map as the view center $VC(x_t,y_t)$, where
\begin{equation}
	(x_t, y_t) = \argmax_{x, y}\big(\mathcal{H}_{on}(t)+\mathcal{H}_{off}(t)\big).
\end{equation}
The positional error between the previously detected target location $(\hat{x}_t, \hat{y}_t)$ and the current view center is computed as
\begin{equation}
	E(x_t,y_t) = \big(\hat{x}_t - VC(x_t), \hat{y}_t - VC(y_t)\big).
\end{equation}

We define a nonlinear transfer function that converts the positional error into a signed position response. 
The exponential term amplifies larger errors, while the sign preserves the direction of the error. 
The horizontal position response is computed as
\begin{equation}
	\begin{aligned}
		\mathcal{S}(x_t) = \operatorname{\mathbf{sgn}}\big(E(x_t)\big) \cdot \left(e^{\left(\frac{4\cdot E(x_t)}{K}\right)^2} - 1\right).
	\end{aligned}
\end{equation}
The vertical position response can be obtained in the same way as $\mathcal{S}(y_t)$. 
Let vector $\mathbf{v}=(\mathcal{S}(x_t),\mathcal{S}(y_t))$, the position response is taking its $\mathcal{L}_2$-norm as $\mathcal{S}(t)=\| \mathbf{v} \|_2$.

Regarding the local motion response, it is calculated by local T4 and T5 interneuron response as 
\begin{equation}
	\mathcal{M}(t) = \big(T4_r(x,y,t) + T5_r(x,y,t) - T4_l(x,y,t) - T5_l(x,y,t)\big).
\end{equation}

Subsequently, we fuse the motion and position responses producing the steering response as
\begin{equation}
	\begin{aligned}
		\Gamma(t) = g_m \cdot \mathcal{M}(t) + g_p \cdot \mathcal{S}(t).
		\label{Eq:sr}
	\end{aligned}
\end{equation}
The steering response is first temporally delayed using a first-order low-pass filter. 
The sign of the delayed response is then compared with that of the previous response. 
The delayed response is accepted only when the two responses have the same sign; otherwise, the original response is retained. 
That is,
\begin{equation}
	\hat{\Gamma}(t) = 
	\begin{cases} 
		\alpha \Gamma(t) + (1-\alpha) \hat{\Gamma}(t-\Delta_t), & \Gamma(t) \cdot \hat{\Gamma}(t-\Delta_t) \geq 0, \\
		\Gamma(t), & \text{otherwise},
	\end{cases}
\end{equation}
where $\alpha \in (0,1)$. 
Accordingly, the visual-field center is updated according to the steering responses along the horizontal (abscissa) and vertical (ordinate) axes.

\begin{figure*}[t]  
	\centering
	\vspace{-20pt}
	\includegraphics[width=0.92\textwidth]{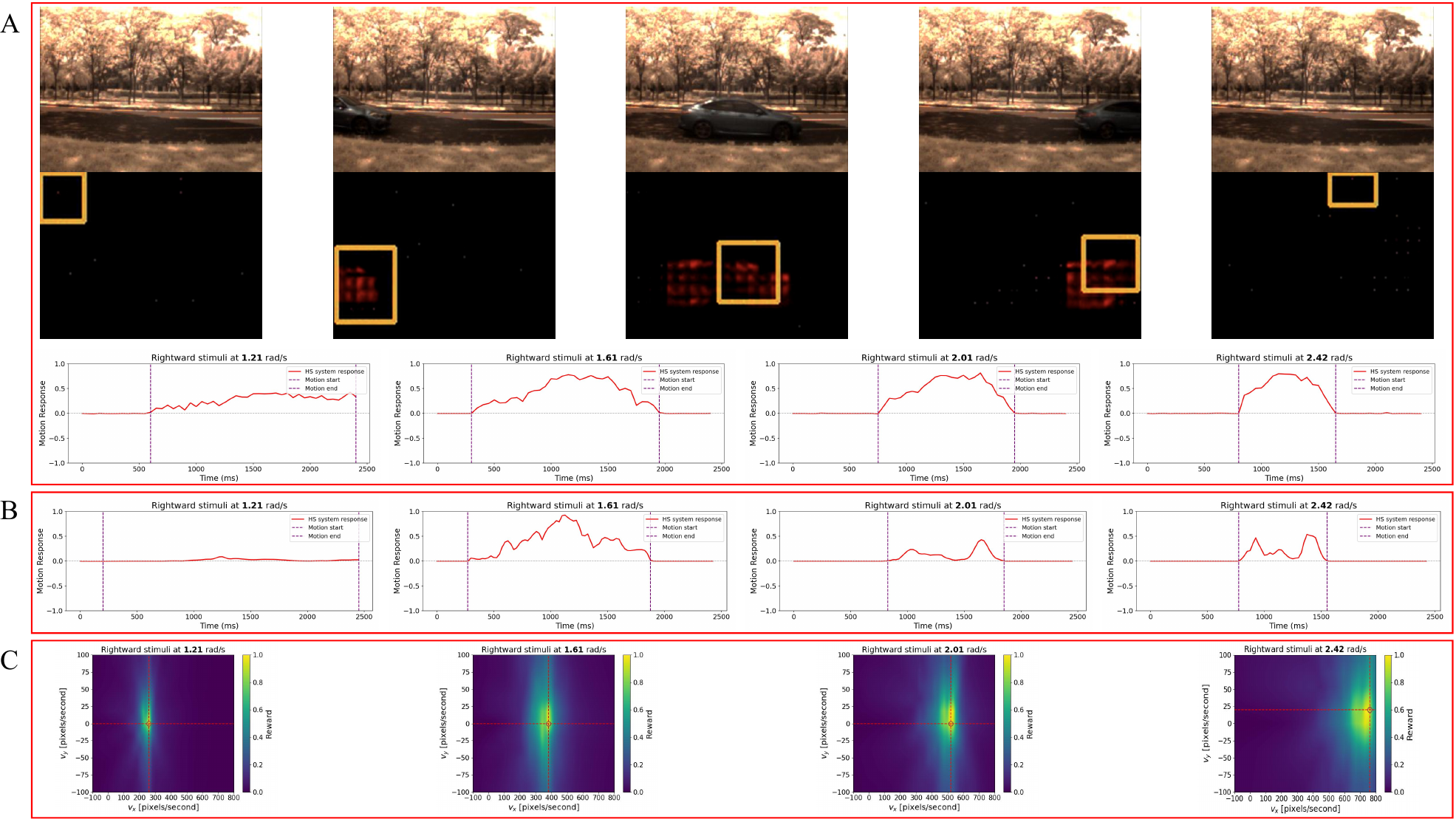}
	\caption{
		Results under daylight conditions. 
		Panel A presents the APS intensity sample frames (first row), the attention saliency maps generated by the proposed model (second row), and the corresponding motion responses indicating motion direction and magnitude at different visual speeds (third row). 
		Panel B shows the motion responses of the frame-based model computed from the APS intensity frames. 
		Panel C presents the image velocities estimated from the input event streams using the contrast-maximization method.		
		}
	\label{fig:EX1}
	\vspace{-10pt}
\end{figure*}

Finally, inspired by the selective-attention behavior of flying insects, the updated visual-field center is treated as the focus of attention. 
Starting from this center, a connected-region expansion process is performed to generate an adaptive attention area. 
Specifically, we define a time-varying region of interest, $\Omega(t)$, within the visual field, with a maximum search side length $2R$. 
Unlike a fixed-size attention window, the proposed attention region adaptively adjusts to the spatial extent of the target, thereby enclosing the motion boundaries of the foreground object within a unified region. 
That is, 
\begin{equation}
	\Omega(t) = \{ (x_t, y_t) \mid |x_t - VC(x_t)| < R, |y_t - VC(y_t)| < R \}.
	\label{Eq:roi}
\end{equation}

\begin{figure*}[t]                
	\centering
	\vspace{-20pt}
	\includegraphics[width=0.92\textwidth]{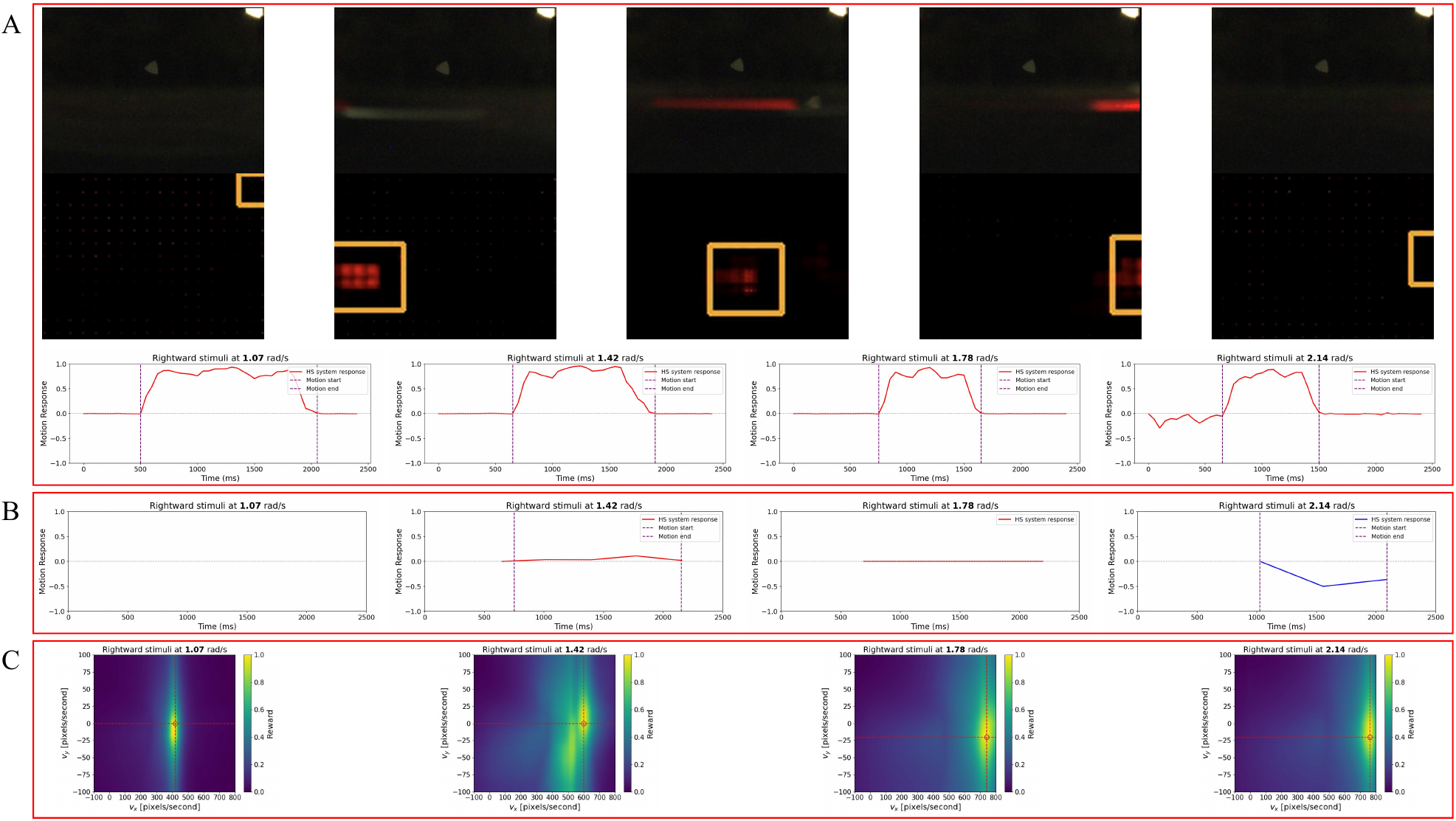}
	\caption{
		Results under nighttime conditions. 
		All notations are consistent with those in Fig.~\ref{fig:EX1}. 
		Owing to hardware limitations, the APS frame rate decreases substantially at night, falling below 10~fps and, in some cases, failing to produce valid intensity frames.
	}
	\vspace{-10pt}
	\label{fig:EX2}
\end{figure*}

For the sake of computing the motion response along four cardinal directions $\mathcal{D}$, the outputs of T4 and T5 interneurons are integrated by the lobula plate tangential cells (LPTC), and multiplied by the binary region of interest $\Omega(t)$. 
Taking the rightward-direction motion computation as the example, 
that is, 
\begin{equation}
		LP_r(t) = \sum_{x=1}^{K} \sum_{y=1}^{K} \mathcal{A}(x,y,t) \cdot \big( T4_r(x,y,t) + T5_r(x,y,t) \big),
\end{equation}
where 
\begin{equation}
	\mathcal{A}(x,y,t) = 
	\begin{cases} 
		1, & \text{if } (x, y) \in \Omega(t) \\
		0, & \text{otherwise}
	\end{cases}.
\end{equation}

The LPTC forms two motion-sensitive output systems of the proposed neural model as
\begin{equation}
	\begin{aligned}
		HS(t) &= LP_r(t) - LP_l(t),\\
		VS(t) &= LP_d(t) - LP_u(t),
	\end{aligned}
\end{equation}
where positive and negative responses in the horizontal-system (HS) pathway indicate preferred-direction rightward and null-direction leftward motion, respectively. 
Similarly, in the vertical-system (VS) pathway, positive and negative responses indicate downward and upward motion, respectively. 
The resulting preferred-direction and null-direction are normalized into $[0,1)$ and $(-1,0]$, respectively. 
We herein employ a winner-take-all mechanism to output either the HS or the VS system motion response.

\subsection{Setting the Parameters}

The proposed computational framework involves only a small number of adjustable parameters, as summarized in Table~\ref{Tab-Parameters}. 
Since the framework is derived from previously well-established methods, no learning or training procedures were employed in this study.

\section{Experimental Evaluation}
\label{Sec: experiments}

This section presents the experimental evaluation of the proposed method, including comparisons with the frame-based fly-inspired network \cite{fu2020} and the event-driven contrast-maximization method \cite{contrast2019}.

All experiments were conducted on a Mac platform equipped with an Apple M4 chip featuring a 10-core CPU and a 10-core GPU, together with 16~GB of RAM and a 512~GB solid-state drive. 
The datasets were collected using a DAVIS346 event camera under both daylight and nighttime conditions. 
Four vehicle-motion sequences were recorded at four different driving velocities in each condition. 
Specifically, the vehicle moved rightward across the camera's field of view at measured angular velocities of 1.21, 1.61, 2.01, and 2.42~rad/s in the daytime road scene, and 1.07, 1.42, 1.78, and 2.14~rad/s in the nighttime road scene. 
Meanwhile, APS intensity frames were synchronously captured by the DAVIS346 camera and used as input to the frame-based comparative model.

Fig.~\ref{fig:EX1} and~\ref{fig:EX2} compare the performance of the evaluated models under daytime and nighttime visual-motion stimuli. 
The proposed event-based framework produces clearer and more stable motion responses than the frame-based model. 
In contrast, the frame-based responses are generally weaker and more variable, particularly at low and high motion speeds. 
At the lowest angular velocity of (1.21~rad/s), the frame-based input produces only a weak response, whereas the event-based representation still yields clearly distinguishable motion activity. 
As the angular velocity increases, the proposed method maintains consistent response characteristics, while the frame-based responses become less smooth and exhibit multiple local peaks. 
Under nighttime conditions, the frame-based method \cite{fu2020} largely loses its motion-detection capability, whereas the proposed method continues to provide stable responses. 
These results indicate that asynchronous event streams preserve motion information more effectively than conventional intensity frames, particularly under challenging illumination conditions.

\begin{figure}[t]
	\centering
	\includegraphics[width=0.5\textwidth]{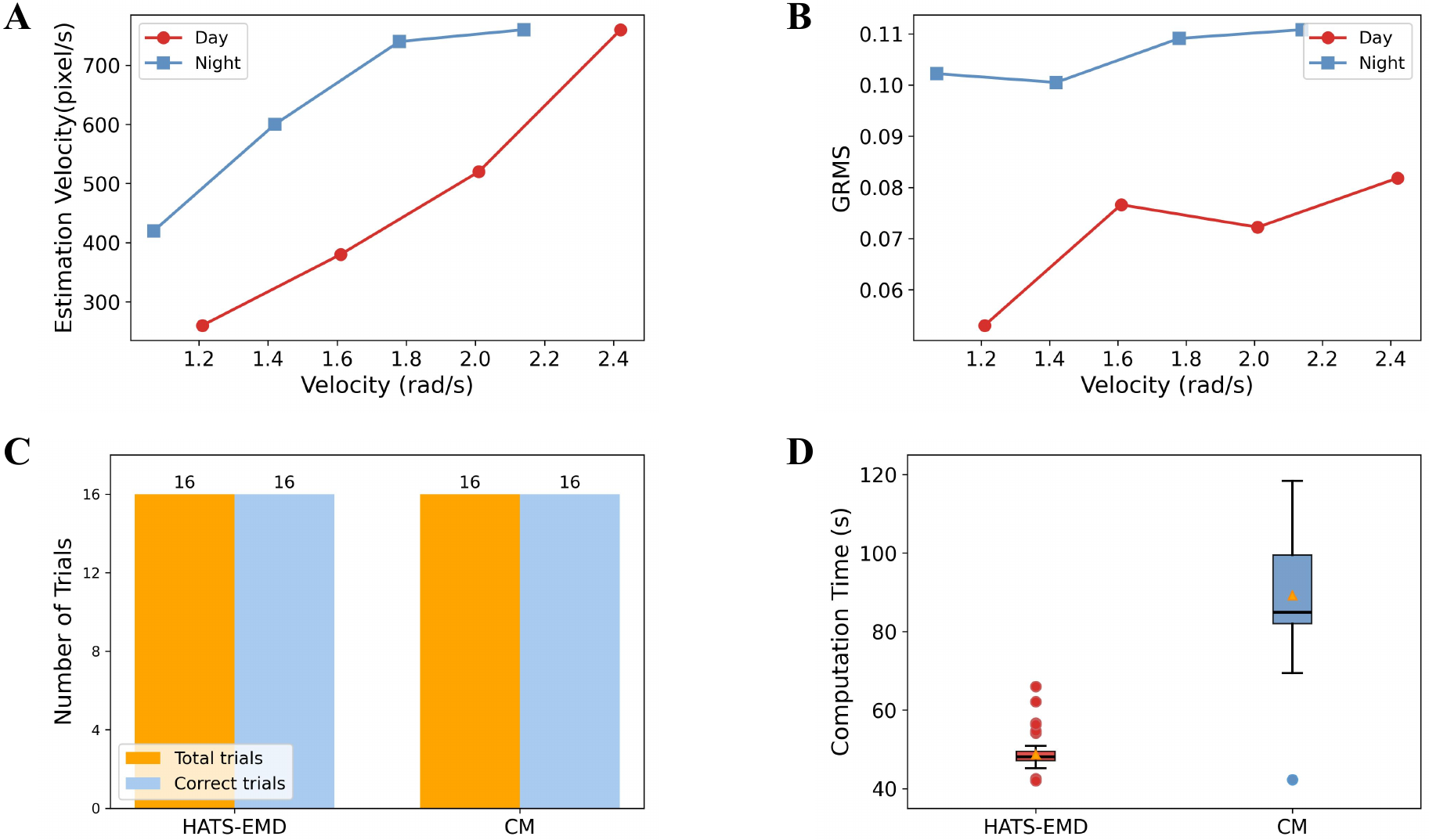}
	\caption{
		Systematic evaluation results. 
		Panel A presents the image-velocity estimates obtained using the comparative contrast-maximization (CM) method. 
		Panel B shows the gradient root mean square (GRMS) responses of the proposed method under different motion velocities. 
		Panel C compares the motion-direction prediction results of the proposed and comparative methods. 
		Panel D summarizes the computational time costs of the evaluated methods.
	}
	\label{fig:EX3}
	\vspace{-10pt}
\end{figure}

In addition, contrast maximization (CM) is an event-based image-velocity estimation method that performs motion compensation by searching for the velocity parameters that maximize the contrast of an event image warped to a reference time \cite{contrast2019}. 
Fig.~\ref{fig:EX1} and~\ref{fig:EX2} present the CM results obtained on our dataset. 
Both the proposed method and the CM approach consistently identify the dominant rightward motion direction. 
As further shown in Fig.~\ref{fig:EX3}, both methods correctly predict the motion directions across all tested sequences and exhibit sensitivity to motion velocity. 
For the proposed method, this velocity sensitivity is quantified using the gradient root mean square (GRMS), whose response magnitude generally increases with motion velocity. 
Nevertheless, the proposed method achieves this directional motion estimation with a much lower computational time cost than the CM approach. 
This finding highlights the advantage of biologically inspired structured neural processing as a lightweight and efficient approach to visual motion detection.

\section{Concluding Remarks}
\label{Sec: conclusion}

This paper presents an event-driven, bio-inspired framework for visual motion detection that integrates the complementary advantages of event-based sensing and structured neural modeling. 
The proposed method combines time-surface-based event synchronization, fly-optic-lobe-inspired motion-feature extraction in complex and dynamic scenes, and rapid visual enhancement through a bottom-up salience built-up. 
Experimental results validate its effectiveness in low-light scenarios compared to frame-based model, and computational efficiency compared to optimization-based approach.


\begin{thebibliography}{00}
\bibitem{seg2019} T. Stoffregen, G. Gallego, T. Drummond, L. Kleeman, and D. Scaramuzza, “Event-based motion segmentation by motion compensation,” in Proc. Int. Conf. Comput. Vis., 2019, pp. 7243–7252.
\bibitem{max2025} Pritam P. Karmokar, Quan H. Nguyen, and William J. Beksi, "Secrets of Edge-Informed Contrast Maximization for Event-Based Vision." 2025 IEEE/CVF Winter Conference on Applications of Computer Vision (WACV). IEEE, 2025.
\bibitem{track2015}Z. Ni, S. -H. Ieng, C. Posch, S. Regnier, and R. Benosman, “Visual tracking using neuromorphic asynchronous event-based cameras,” Neural Comput., vol. 27, no. 4, pp. 925–953, 2015.
\bibitem{brosch2015} T. Brosch, S. Tschechne, and H. Neumann, “On event-based optical flow detection,” Front. Neurosci., vol. 9, Apr. 2015, Art. no. 137.
\bibitem{orchard2013} G. Orchard, R. Benosman, R. Etienne-Cummings, and N. V. Thakor, “A spiking neural network architecture for visual motion estimation,” in Proc. IEEE Biomed. Circuits Syst. Conf., 2013, pp. 298–301.
\bibitem{IBM2018}G. Haessig, A. Cassidy, R. Alvarez, R. Benosman, and G. Orchard, “Spiking optical flow for event-based sensors using IBM’s True North neurosynaptic system,” IEEE Trans. Biomed. Circuits Syst., vol. 12, no. 4, pp. 860–870, Aug. 2018.
\bibitem{unsuper2020} F. Paredes-Valles, K. Y. W. Scheper, and G. C. H. E. de Croon, “Unsupervised learning of a hierarchical spiking neural network for optical flow estimation: From events to global motion perception,” IEEE Trans. Pattern Anal. Mach. Intell., vol. 42, no. 8, pp. 2051–2064, Aug. 2020.
\bibitem{fu2020} Q. Fu and S. Yue, “Modelling Drosophila motion vision pathways for decoding the direction of translating objects against cluttered moving backgrounds,” Biol. Cybern., vol. 114, no. 4, p. 443. 2020.
\bibitem{contrast2019} T. Stoffregen and L. Kleeman, “Event cameras, contrast maximization and reward functions: An analysis,” in Proc. IEEE/CVF Conf. Comput. Vis. Pattern Recognit. (CVPR), 2019.
\bibitem{ziyan2025} Z. Qin, J. Peng, S. Yue and Q. Fu, "A bio-inspired research paradigm of collision perception neurons enabling neuro-robotic integration: the LGMD case," Journal of the Royal Society Interface 22:20250433, 2025.
\bibitem{fu2026} Q. Fu and Z. Qin, "Emergence of robust looming selectivity via coordinated inhibitory neural computations," Applied Soft Computing, 187:114337, 2026.
\bibitem{event-driving-2020} G. Chen, H. Cao, J. Conradt, H. Tang, F. Rohrbein and A. Knoll, "Event-based neuromorphic vision for autonomous driving: A paradigm shift for bio-inspired visual sensing and perception," IEEE Signal Processing Magazine, 37(4), 34-49, 2020.
\bibitem{event-survey} G. Guillermo, D. Tobi, O. Garrick, B. Chiara, T. Brian, C. Andrea, L. Stefan, J. D. Andrew, C. Jorg, D. Kostas, S. Davide, "Event-based Vision: A Survey," IEEE Transactions on Pattern Analysis and Machine Intelligence, 44(1), 154-180, 2020.
\end{thebibliography}
\end{document}